\documentclass{article}

\usepackage{PRIMEarxiv}

\usepackage[utf8]{inputenc} 
\usepackage[T1]{fontenc}    
\usepackage{hyperref}       
\usepackage{url}            
\usepackage{booktabs}       
\usepackage{amsfonts}       
\usepackage{nicefrac}       
\usepackage{microtype}      
\usepackage{lipsum}
\usepackage{fancyhdr}       
\usepackage{graphicx}       
\usepackage{multirow}
\usepackage{gensymb}

\title{3D-CNN for Facial Micro- and Macro-expression Spotting on Long Video Sequences using Temporal Oriented Reference Frame
}

\author{
  Chuin Hong Yap \\
  Manchester Metropolitan University \\
  Manchester, UK \\
  \texttt{chuin.h.yap@stu.mmu.ac.uk} \\
   \And
  Moi Hoon Yap \\
    Manchester Metropolitan University \\
  Manchester, UK \\
  \texttt{M.Yap@mmu.ac.uk} \\
   \AND
   Adrian K. Davison \\
   University of Manchester \\
   Manchester, UK \\
   \And
   Connah Kendrick \\
   Manchester Metropolitan University \\
   Manchester, UK \\
   \AND
   Jingting Li, Sujing Wang \\
    CAS Key Laboratory of Behavioral Science, \\
    Institute of Psychology \\
   Beijing, China \\
   \And
   Ryan Cunningham \\
Manchester Metropolitan University \\
  Manchester, UK \\
  \texttt{R.Cunningham@mmu.ac.uk} \\
}

\begin{document}
\maketitle

\begin{abstract}
Facial expression spotting is the preliminary step for micro- and macro-expression analysis. The task of reliably spotting such expressions in video sequences is currently unsolved. The current best systems depend upon optical flow methods to extract regional motion features, before categorisation of that motion into a specific class of facial movement. Optical flow is susceptible to drift error, which introduces a serious problem for motions with long-term dependencies, such as high frame-rate macro-expression. We propose a purely deep learning solution which, rather than tracking frame differential motion, compares via a convolutional model, each frame with two temporally local reference frames. Reference frames are sampled according to calculated micro- and macro-expression duration. As baseline for MEGC2021 using leave-one-subject-out evaluation method, we show that our solution achieves F1-score of 0.105 in a high frame-rate (200 fps) SAMM long videos dataset (SAMM-LV) and is competitive in a low frame-rate (30 fps) (CAS(ME)\textsuperscript{2}) dataset. On unseen MEGC2022 challenge dataset, the baseline results are 0.1176 on SAMM Challenge dataset, 0.1739 on CAS(ME)\textsuperscript{3} and overall performance of 0.1531 on both dataset. 
\end{abstract}


\section{Introduction}
\label{sec:introduction}

Facial expression is the main way people convey visual information of human emotion. It can predict a person's current state of emotion.
Facial expressions can be classified into two groups: macro-expression (MaE) and micro-expression (ME). These classifications are based on their relative duration and intensity, where MaE (also known as a regular facial expression) lasts from 0.5 to 4.0s \cite{yan2013fast} and has higher intensity; ME occurs in less than 0.5s and has lower intensity. ME occurs more frequently in high-stake and stressful circumstances \cite{ekman2003darwin, ekman2004emotions}. As it is an involuntary reaction, the emotional state of a person can be revealed through analysing MEs.

Earlier works of ME are based on datasets of short clips containing categorised ME (i.e., SAMM \cite{davison2018samm, davison2018objective}, SMIC \cite{li2013spontaneous}, and CASME II \cite{yan2014casme}). These were used to facilitate ME expressions recognition \cite{yap2018facial, see2019megc}. With recent interest in ME and MaE spotting, researchers created long video datasets, SAMM Long Videos (SAMM-LV) \cite{yap2020samm} and CAS(ME)\textsuperscript{2} \cite{qu2017cas}, to better represent spontaneous emotion for ME and MaE spotting.
This paper focuses on automated spotting of MaE and ME on SAMM-LV and CAS(ME)\textsuperscript{2}. We produce the baseline results for two Facial Micro-expressions Grand Challenge (MEGC), i.e. MEGC2021 \cite{li2021fme} and MEGC2022. To increase the level of challenge, we introduce a new unseen dataset.  

Most of the previous methods utilise long short-term memory (LSTM) \cite{verburg2019micro, chanti2019ads} or optical flow \cite{verburg2019micro, sun2019two, he2020spotting, zhang2020spatio} to detect temporal correlation of video sequences. LSTM is a recurrent neural network that computes sequential time steps with a new element of the input sequence being added to the network at each time step \cite{sen2018approximate}.
Optical flow computes the differences of two image frames every time when it is applied within a video sequence.
Both LSTM and optical flow are computationally expensive.
In addition, optical flow has weaknesses such as drifting over frames \cite{bertero1988ill} and is very susceptible to illumination changes \cite{turaga2010advances}.
We also noticed that previous attempts lack duration centred analysis. We take advantage of the major difference between ME and MaE (they occur for different duration, where ME occurs less than 0.5s while MaE occurs in 0.5s or longer) and propose a two-stream network with a different frame skip based on the duration differences for ME and MaE spotting.

The main contributions are:
\begin{itemize}
    \item Our approach is the first end-to-end deep learning ME and MaE spotting method trained from scratch using long video datasets.
    \item Our method uses a two-stream network with temporal oriented reference frame. The reference frames are two frame pairs corresponding to the duration difference of ME and MaE. The two-stream network also possesses shared weights to mitigate overfitting.
    \item The network architecture consists of only 3 convolutional layers with the capability of detecting co-occurrence of ME and MaE using a multi-label system. This method has the potential to be used on lightweight devices (e.g., smartphones) in real-time.
    \item To make the network less susceptible to uneven illuminations, Local Contrast Normalisation (LCN) is included into our network architecture. LCN drastically improves the overall network performance across a range of configurations and parameters. 
\end{itemize}

\begin{figure*}
	\begin{center}
		\includegraphics[width=.95\linewidth]{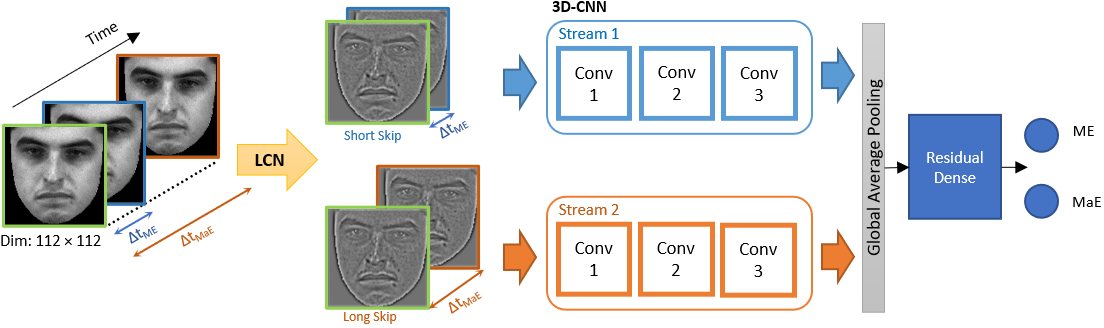}
	\end{center}
	\caption{Network architecture of our two-stream 3D-CNN. Our network has only 3 layers (4 layers included LCN). Temporal oriented frame skip based on the duration differences of ME and MaE (where $\Delta t_{ME} < \Delta t_{MaE}$). LCN is applied using a convolutions kernel which performs local contrast normalisation as described in Equation \ref{eq:LCN_eqn}. Each convolutional block consists of depthwise separable convolution, batch normalisation and dropout.
	The residual dense layer possesses the skip connections that shares weights. Two dense nodes were used at the end to resemble the presence of ME and MaE.}
	\label{fig:network architecture}
\end{figure*}

\section{Proposed Method}
Our goal is to detect ME and MaE within long video sequences. By using the duration difference of ME and MaE, we propose a two-stream 3D-Convolutional Neural Network (3D-CNN) with temporal oriented frame skips. We define the two ``streams" as ME and MaE pathways, as illustrated in Fig.~\ref{fig:network architecture}. They are structurally identical networks with shared weights, but differ in frame skips. 
We use 3 convolutional layers and pool all the spatial dimensions before the dense layers using global average pooling. This design constrains the network to focus on regional features, rather than global facial features. Next, we further propose that normalising the brightness and/or contrast of the images. This is important for generalisation and real world applications, as there is likely more variation in skin tone and brightness between different individuals, and lighting conditions. Therefore, we apply LCN to all images before presented to our network.

\subsection{Preprocessing}
\label{sec:preprocessing}

\textbf{Facial Alignment} OpenFace 2.0 \cite{baltrusaitis2018openface} is used for facial alignment. It is a general-purpose toolbox for facial analysis.
OpenFace uses Convolutional Experts Constrained Local Model (CE-CLM) \cite{zadeh2017convolutional} of 84-points for facial landmark tracking and detection.
Based on the detected facial landmarks, the face in each frame of a video sequence is aligned and extracted.
In our experiment, image resolution is 112$\times$112 pixels, which is the default output resolution of OpenFace.

\begin{figure}
	\begin{center}
		\includegraphics[width=.6\linewidth]{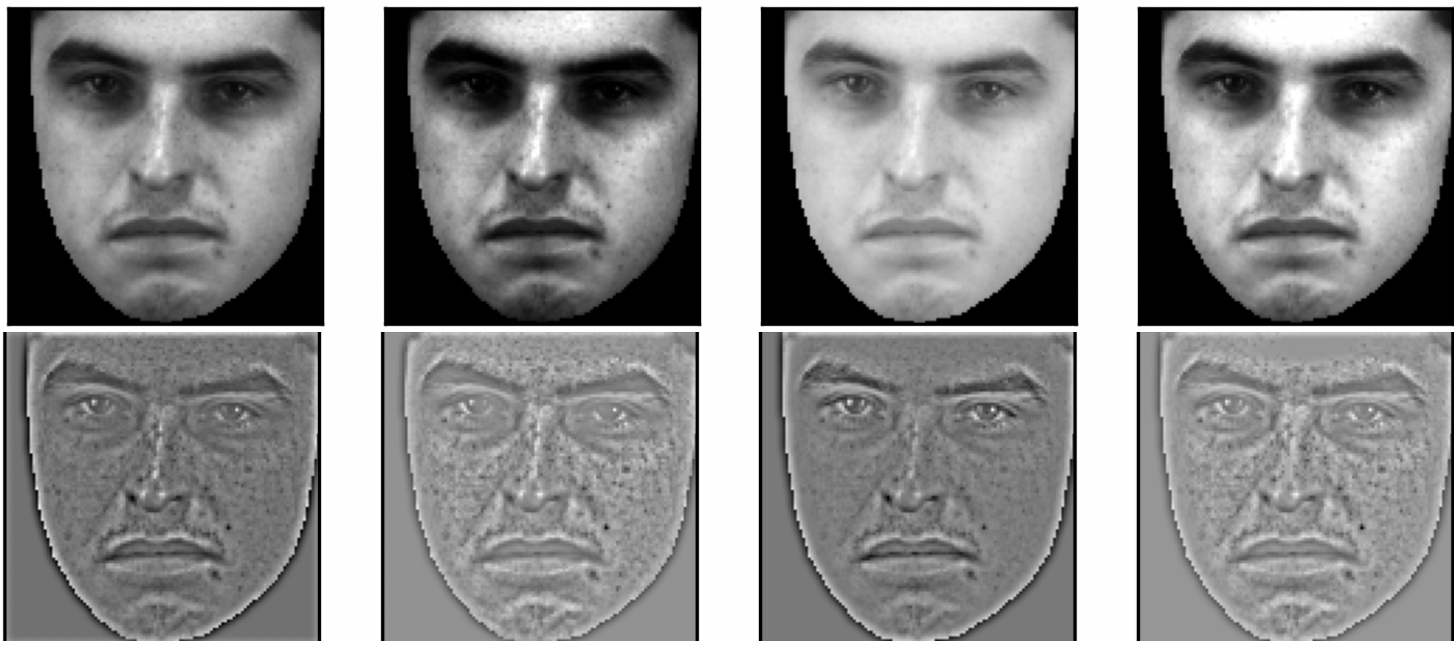}
	\end{center}
	\caption{Preprocessing: (Top) Face alignment and data augmentation (randomised brightness and contrast change) on a subject of SAMM-LV; and (Bottom) Image normalised using LCN. Despite the brightness and contrast differences, the facial features remain well-preserved.}
	\label{fig:LCN samm-lv}
\end{figure}

\noindent
\textbf{Local Contrast Normalisation (LCN)} LCN \cite{jarrett2009best} was inspired by computational neuroscience models that mimic human visual perception \cite{lyu2008nonlinear} by mainly enhancing low contrast regions of images.
LCN normalises the contrast of an image by conducting local subtractive and divisive normalisations \cite{jarrett2009best}. It performs normalisation on local patches (per pixel basis) by comparing a central pixel value with its neighbours. The unique feature of LCN is its divisive normalisation, which consists of the maximum of local variance or the mean of global variance. If an area of image has very low variance (approximately 0), dividing with a small value will form a bright spot. Dividing using the mean of global variance mitigates this issue.
The main advantage of this method is robustness towards the change in brightness or contrast (shown in Figure \ref{fig:LCN samm-lv}). The facial features are well preserved despite the random changes in brightness and contrast. This can be a solution to address the weakness of overused conventional optical flow method of dealing with uneven lighting.
In our implementation, Gaussian convolutions are used to obtain the local mean and standard deviation.
Gaussian convolution acts as a low pass filter which reduces noise. It also speeds up the local normalisation process as it is a separable filter (where 2-dimensional data can be calculated using 2 independent 1-dimensional functions).

The general equation of LCN can be described as
\begin{equation} 
	\footnotesize
	g(x,y) = \frac{f(x,y) - m_f(x,y)}{ max\textbf{(} \sigma_f(x,y), c \textbf{)}} \label{eq:LCN_eqn}
\end{equation}

\noindent where $f(x,y)$ is the input image, $m_f(x,y)$ is the local mean estimation, $\sigma_f(x,y)$ is the local variance estimation, $c$ is the mean of local variance estimation and $g(x,y)$ is the output image.

\subsection{Network Architecture}
\label{sec:network architecture}
We propose a two-stream network using a 3D-CNN (network architecture shown in Figure \ref{fig:network architecture}). Our network takes advantage of the duration differences of ME and MaE and encouraging one network to be more sensitive to ME and the other to MaE. This is made possible by using a different number of skipped frames in each respective stream (using the maximum duration of a ME, 0.5s, as the threshold for the duration difference). Our network consists of depthwise separable convolutions, which has about 10\% less parameters compared to regular convolution counterpart.

\noindent
\textbf{Input Layer} The input of this network consists of 4 images. The frame pair in the first stream has a shorter frame skip compared to the latter pair. The frame skips are determined based on the $k$-th frame. The $k$-th frame, described by Moilanen et al. \cite{moilanen2014spotting}, is the average mid-point of odd-numbered facial expression interval of the whole dataset. These pairs are then fed into two separate but identical neural networks with shared weights.

\noindent
\textbf{Weighted loss function} To the best of our knowledge, we are the first in ME spotting to weight imbalanced datasets using a loss function. The datasets used in our experiment are imbalanced, and there are more neutral frames relative to frames containing ME or MaE. 
We also weighted the loss based on ME and MaE, as ME occurs less than MaE. The loss can be described as
\begin{equation} \label{eq: weighted loss function movement}
    \footnotesize
    Loss = -\sum\limits_{i=1}^{C'} M_i \cdot [W \cdot t_i \cdot log(s_i)-(1-t_i) \cdot log(1-s_i)]
\end{equation}

\noindent where $t_i$ is ground truth labels, $s_i$ are the predictions, $C'$ is the number of expression types ($C'$=2 in our case, for ME and MaE), \textit{W} is the weighting factor that functions to penalise more when the network predicts ME/MaE wrongly as neutral and $M_i$ is the weighting factor for expression (ME or MaE). 

We only apply weighted loss function when training SAMM-LV as we found out model trained with SAMM-LV improves with weighted loss function. The effects in CAS(ME)\textsuperscript{2} is negligible.
We used $C'$ = 2, $M_0$ = 0.9 (for ME), $M_1$ = 0.1 (for MaE). Coefficient \textit{W} used is 3.
All the weighting factors are used to address the dataset imbalance. \textit{W} is used to address different number of ground truth labels of ME/MaE and neutral; $M_0$ and $M_1$ is used to address the imbalanced labels of ME and MaE.

\noindent
\textbf{Depthwise Separable Convolution} We use depthwise separable convolution of MobileNet \cite{howard2017mobilenets} that reduces total trainable parameters with minimal performance impact. It consists of depthwise and pointwise convolution. Depthwise convolution is convolution applied on individual channels instead of all channel at once (as in regular convolutional). Pointwise convolution is convolution that uses a $1\times1$ kernel with a third dimension of $d$ (where $d$ is the number of channels) on the feature maps.

\noindent
\textbf{GAP and Residual Dense Layer} A global average pooling (GAP) layer is used to flatten the convolution output and enforce modelling of localised facial movements. It is followed by the final hidden layer consisting of a residual dense layer. This layer consists of two fully connected layers with skip connections inspired by ResNet \cite{he2016deep}. 

\noindent
\textbf{Output Layer} The output layer consists of two dense nodes with sigmoid activation representing the presence of ME and MaE. 

\section{Experiment}
\label{sec:experiment}

\noindent
\textbf{MEGC2021 Spotting Datasets.}
The datasets used are SAMM Long Videos (SAMM-LV) \cite{yap2020samm} with 147 long videos containing 343 MaEs and 159 MEs; and CAS(ME)\textsuperscript{2} \cite{qu2017cas} with 87 long videos containing 300 MaEs and 57 MEs. The original ground truth of these datasets consist of onset, apex, and offset frame labels of each facial expression. We label the ground truth of movement from the onset frame to the offset frame, inclusively. Our ground truth consists of two labels of binaries where ``0" represents absence while ``1" represents presence of ME or/and MaE.


\noindent
\textbf{MEGC2022 Unseen Test Set.}
In MEGC2022, we introduce an unseen test set with 10 long video, which consists of 5 long videos from SAMM \cite{davison2018samm} (SAMM Challenge dataset) and 5 clips cropped from different videos in CAS(ME)\textsuperscript{3}~\cite{li2022casme3}. The frame rate for SAMM Challenge dataset is 200 fps and the frame rate for CAS(ME)\textsuperscript{3} is 30 fps. The participants can use SAMM-LV and CAS(ME)\textsuperscript{2} as training set, and test on this unseen dataset. For facilitate the spotting challenge and to enable fair assessment, we do not release the ground truth for this dataset. The participants will submit their results to our grand challenge system (https://megc2022.grand-challenge.org).

\begin{table}[h]
	\begin{center}
		\caption{Training configuration. Stream 1 is designed to be more sensitive to ME, while Stream 2 is more sensitive to MaE by using different range of frame skips based on the duration differences of ME and MaE.
	The $k$-th frame is the average mid-point of facial expression interval.
	Note: $^{\star}$ used in training and validation, $^{\dagger}$ used in testing
	}
       \scriptsize
		\renewcommand{\arraystretch}{1}
		\begin{tabular}{|l|c|c|}
			\hline
			\textbf{Dataset} & \textbf{SAMM-LV} & \textbf{CAS(ME)\textsuperscript{2}} \\
			\hline
			Random frame skip$^{\star}$ (Stream 1 \& 2) &25$\sim$75 \& 200$\sim$400 &3$\sim$9 \& 16$\sim$50\\
			\hline
            $k$-th frame skip$^{\dagger}$ (Stream 1 \& 2) &37 \& 217 &6 \& 19\\
			\hline
			Manual frame skip$^{\dagger}$ (Stream 1 \& 2) &30 \& 310 &10 \& 33\\
			\hline
		\end{tabular}
		\label{tab:training configuration}
	\end{center}
\end{table}
\noindent
\textbf{Training.}
Randomised frame skips are used in training and validation. This creates a more realistic scenario as the duration of each facial expression is unknown in real life. 
For model testing, we used a frame skip based on the $k$-th frame of ME and MaE of each respective dataset shown in Table \ref{tab:training configuration}.
The visual differences of frames calculated using this interval (frames skipped) is larger, making the facial movements more distinct for the algorithm to spot.

\noindent
\textit{Regularisation} Random augmentations (i.e., contrast, gamma intensity, and gamma gain) on the input images are performed with a range of 0.5 to 1.5. Other augmentations include 50\% probability of horizontal flip and $\pm$10$^{\circ}$ of image rotation. Other regularisations include adding dropout layers and random frame skips during training and validation.

\noindent
\textit{Training Configuration} As shown in Table \ref{tab:training configuration}, the results are evaluated using leave-one-subject-out (LOSO) cross-validation. 

\section{Results}
\label{sec:results and discussion}
We apply the Intersection over Union (IoU) method used in Micro-Expression Grand Challenge (MEGC) III \cite{he2020spotting,li2020megc2020} to compare with other methods.  
The interval is then evaluated using the following IoU method
\begin{equation} \label{eq:iou}
    \footnotesize
	\frac{Predicted \cap GT}{Predicted \cup GT} \geq J 
\end{equation}
where $J$ is the minimum overlapping to be classified as true positive, $GT$ represents the ground truth expression interval (onset-offset), $Predicted$ represents the detected expression interval. In our experiment, $J$ is set to 0.5.

As other methods use different post-processing steps, we decided to use two different evaluation methods. The first method is our Automated IoU Method and the second method is Multi-Scale Filter used by Zhang et al. \cite{zhang2020spatio}.

\subsection{Baseline Result for MEGC2021 Spotting Task}
We convert our results into intervals using automated thresholding based on ROC evaluation.
First, the test results are normalised and smoothed using a Butterworth filter \cite{butterworth1930theory}, which is a low-pass filter that cuts off high frequency noises while retaining low frequency signals. The main advantage of this filter is it has a flat magnitude filter whereby signals with frequency below cut-off frequency do not undergo attenuation.
Next, the onset and offset of both ground truth and the predictions are obtained. Finally, the overlapping was analysed using the IoU method (where TP must fulfill the criteria in Equation \ref{eq:iou}).

\begin{table}[h]
	\footnotesize
	\caption{F1-score of ME and MaE spotting using our Automated IoU Method, where Ours* represents our proposed method with $k$-th frame skip and Ours** represents our proposed method with manual frame skip. Manual frame skip is performed by first taking $k$-th frame as a reference, proceeded by increasing or decreasing the frame skips until the results improve.} 
	\centering
	\renewcommand{\arraystretch}{1}       
	\begin{tabular}{ | c || c | c | c | c | c | c | }	\hline
		\multicolumn{1}{|c||}{\multirow{2}{*}{\textbf{Method}}} & \multicolumn{3}{c|}{\textbf{SAMM-LV}} & \multicolumn{3}{c|}{\textbf{CAS(ME)\textsuperscript{2}}} \\
		\cline{2-7}
		 & MaE & ME & Overall & MaE & ME & Overall \\
		\hline
		Pan \cite{pan2020local}  & -        & -    & 0.0813	& -	& -	& 0.0595      \\ \hline
		\textbf{Ours*}          & 0.1504          & 0.0421          & 0.1017          & 0.0704          & 0.0075 & 0.0509          \\ \hline
\textbf{Ours**}         & \textbf{0.1543} & \textbf{0.0442} & \textbf{0.1050} & \textbf{0.0874} & 0.0075          & \textbf{0.0630} \\ \hline	
	\end{tabular}
	\label{tab:f1-score MaE and ME spotting task}
\end{table}

Our results show better spotting performance in SAMM-LV compared to CAS(ME)\textsuperscript{2}. One possibility is SAMM-LV has higher frame rate (200 fps) and the randomised frame skipping used in our training pipeline has more variety of input data to be learnt compared to CAS(ME)\textsuperscript{2} (30 fps). Hence, our model is able to learn data with more variation in SAMM-LV and show better performance. ME which occur in less than 0.5s, has a small window of detection. A lower ME detection rate in CAS(ME)\textsuperscript{2} might also be a consequence of the lower frame rate.

Zhang et al. \cite{zhang2020spatio} and He et al. \cite{he2020spotting} are conventional approaches. These methods use post-processing steps to enhance ME spotting rate. Hence, it is not fair to compare our method directly. Instead, we use Zhang et al.'s post-processing steps (also named Multi-Scale Filter \cite{zhang2020spatio}) and the results are shown in Table \ref{tab:f1-score MaE and ME spotting task manual}.
We obtained a notable improvement in ME and MaE spotting, particularly in CAS(ME)\textsuperscript{2}. By implementing these post-processing steps, our method outperforms in SAMM-LV and CAS(ME)\textsuperscript{2} in MaE spotting and overall performance. Although we obtained better results using this evaluation, 
we noticed that this method requires selection of hyperparameters (e.g., window size and order of Savitzky-Golay filter, the upper limit of interval distance to merge etc).

\begin{table}[h]
	\footnotesize
	\caption{F1-score of ME and MaE spotting using Multi-Scale Filter (manual post-processing steps used by Zhang et al. \cite{zhang2020spatio}), where Ours* represents our proposed method with $k$-th frame skip and Ours** represents our proposed method with manual frame skip. This post-processing steps involves signal smoothing using Savitzky-Golay filter and signal merging when intervals are close to each other.} 
	\centering
	\renewcommand{\arraystretch}{1}       
	\begin{tabular}{ | c || c | c | c | c | c | c | }	\hline
		\multicolumn{1}{|c||}{\multirow{2}{*}{\textbf{Method}}} & \multicolumn{3}{c|}{\textbf{SAMM-LV}} & \multicolumn{3}{c|}{\textbf{CAS(ME)\textsuperscript{2}}} \\
		\cline{2-7}
		 & MaE & ME & Overall & MaE & ME & Overall \\
		\hline
		He \cite{he2020spotting}  & 0.0629	& 0.0364	& 0.0445 & 0.1196        & 0.0082    & 0.0376	      \\ \hline
		Zhang \cite{zhang2020spatio}        	& 0.0725	& \textbf{0.1331}	& 0.0999  & 0.2131       & 0.0547              & 0.1403     \\ \hline
		\textbf{Ours*}          & 0.1569          & 0.0512          & 0.1083 & 0.1880          & 0.0583          & 0.1449          \\ \hline
\textbf{Ours**}         & \textbf{0.1595} & 0.0466          & \textbf{0.1084}          & \textbf{0.2145} & \textbf{0.0714} & \textbf{0.1675} \\ \hline
	\end{tabular}
	\label{tab:f1-score MaE and ME spotting task manual}
\end{table}


\subsection{Baseline Result for MEGC2022 Spotting Task}
Table \ref{tab:f1-score MaE and ME spotting task FMEGC 2022} shows the baseline result to facilitate MEGC2022 Spotting Task. The unseen test dataset is first introduced in this paper.
\begin{table}[h]
	\footnotesize
	\caption{F1-score of ME and MaE spotting on unseen test set of MEGC2022 that uses SAMM Challenge and CAS(ME)\textsuperscript{3}} 
	\centering
	\renewcommand{\arraystretch}{1}       
	\begin{tabular}{ | c || c | c | c | c | c | c | c |}	\hline
		\multicolumn{1}{|c||}{\multirow{2}{*}{\textbf{Method}}} & \multicolumn{3}{c|}{\textbf{SAMM Challenge}} & \multicolumn{3}{c|}{\textbf{CAS(ME)\textsuperscript{3}}} & \multirow{2}{*}{\textbf{Overall}} \\
		\cline{2-7}
		& MaE & ME & Overall & MaE & ME & Overall & \\
		\hline
		\textbf{Ours*}          & 0.1739          & 0.0714          & 0.1176           & 0.1622          & 0.2222  & 0.1739   & 0.1351    \\ \hline
	\end{tabular}
	\label{tab:f1-score MaE and ME spotting task FMEGC 2022}
\end{table}

\section{Conclusion}
We presented a temporal oriented two-stream 3D-CNN model that shows promising results in ME and MaE spotting in long video sequences. Our method took advantage of the duration difference of ME and MaE by making a two-stream network that is sensitive to each expression type.
Despite only having 3 convolutional layers, our model showed state-of-the-art performance in SAMM-LV and remained competitive in CAS(ME)\textsuperscript{2}. 
LCN has proven to have significant improvement in our model and the ability to address uneven illumination, which is a major weakness of optical flow. 


\bibliographystyle{unsrt}  
\bibliography{references}

\begin{thebibliography}{10}

\bibitem{yan2013fast}
Wen-Jing Yan, Qi~Wu, Jing Liang, Yu-Hsin Chen, and Xiaolan Fu.
\newblock How fast are the leaked facial expressions: The duration of
  micro-expressions.
\newblock {\em Journal of Nonverbal Behavior}, 37(4):217--230, 2013.

\bibitem{ekman2003darwin}
Paul Ekman.
\newblock Darwin, deception, and facial expression.
\newblock {\em Annals of the New York Academy of Sciences}, 1000(1):205--221,
  2003.

\bibitem{ekman2004emotions}
Paul Ekman and Gavin Yamey.
\newblock Emotions revealed: recognising facial expressions: in the first of
  two articles on how recognising faces and feelings can help you communicate,
  paul ekman discusses how recognising emotions can benefit you in your
  professional life.
\newblock {\em Student BMJ}, 12:140--142, 2004.

\bibitem{davison2018samm}
A.~K. Davison, C.~Lansley, N.~Costen, K.~Tan, and M.~H. Yap.
\newblock Samm: A spontaneous micro-facial movement dataset.
\newblock {\em IEEE Transactions on Affective Computing}, 9(1):116--129, Jan
  2018.

\bibitem{davison2018objective}
Adrian~K Davison, Walied Merghani, and Moi~Hoon Yap.
\newblock Objective classes for micro-facial expression recognition.
\newblock {\em Journal of Imaging}, 4(10):119, 2018.

\bibitem{li2013spontaneous}
Xiaobai Li, Thorsten Pfister, Xiaohua Huang, Guoying Zhao, and Matti
  Pietikainen.
\newblock A spontaneous micro-expression database: Inducement, collection and
  baseline.
\newblock In {\em Automatic Face and Gesture Recognition (FG), 2013 10th IEEE
  International Conference and Workshops on}, pages 1--6. IEEE, 2013.

\bibitem{yan2014casme}
Wen-Jing Yan, Xiaobai Li, Su-Jing Wang, Guoying Zhao, Yong-Jin Liu, Yu-Hsin
  Chen, and Xiaolan Fu.
\newblock Casme ii: An improved spontaneous micro-expression database and the
  baseline evaluation.
\newblock {\em PloS one}, 9(1), 2014.

\bibitem{yap2018facial}
Moi~Hoon Yap, John See, Xiaopeng Hong, and Su-Jing Wang.
\newblock Facial micro-expressions grand challenge 2018 summary.
\newblock In {\em 2018 13th IEEE International Conference on Automatic Face \&
  Gesture Recognition (FG 2018)}, pages 675--678. IEEE, 2018.

\bibitem{see2019megc}
John See, Moi~Hoon Yap, Jingting Li, Xiaopeng Hong, and Su-Jing Wang.
\newblock Megc 2019--the second facial micro-expressions grand challenge.
\newblock In {\em 2019 14th IEEE International Conference on Automatic Face \&
  Gesture Recognition (FG 2019)}, pages 1--5. IEEE, 2019.

\bibitem{yap2020samm}
Chuin~Hong Yap, Connah Kendrick, and Moi~Hoon Yap.
\newblock Samm long videos: A spontaneous facial micro-and macro-expressions
  dataset.
\newblock In {\em 2020 15th IEEE International Conference on Automatic Face and
  Gesture Recognition (FG 2020) (FG)}, pages 194--199, Los Alamitos, CA, USA,
  may 2020. IEEE Computer Society.

\bibitem{qu2017cas}
Fangbing Qu, Su-Jing Wang, Wen-Jing Yan, He~Li, Shuhang Wu, and Xiaolan Fu.
\newblock Cas (me)\^{} 2: A database for spontaneous macro-expression and
  micro-expression spotting and recognition.
\newblock {\em IEEE Transactions on Affective Computing}, 2017.

\bibitem{li2021fme}
Jingting Li, Moi~Hoon Yap, Wen-Huang Cheng, John See, Xiaopeng Hong, Xiaobai
  Li, and Su-Jing Wang.
\newblock {\em FME'21: 1st Workshop on Facial Micro-Expression: Advanced
  Techniques for Facial Expressions Generation and Spotting}, page 5700–5701.
\newblock Association for Computing Machinery, New York, NY, USA, 2021.

\bibitem{verburg2019micro}
Michiel Verburg and Vlado Menkovski.
\newblock Micro-expression detection in long videos using optical flow and
  recurrent neural networks.
\newblock In {\em 2019 14th IEEE International Conference on Automatic Face \&
  Gesture Recognition (FG 2019)}, pages 1--6. IEEE, 2019.

\bibitem{chanti2019ads}
Dawood~Al Chanti and Alice Caplier.
\newblock Ads-me: Anomaly detection system for micro-expression spotting.
\newblock {\em arXiv preprint arXiv:1903.04354}, 2019.

\bibitem{sun2019two}
Bo~Sun, Siming Cao, Jun He, and Lejun Yu.
\newblock Two-stream attention-aware network for spontaneous micro-expression
  movement spotting.
\newblock In {\em 2019 IEEE 10th International Conference on Software
  Engineering and Service Science (ICSESS)}, pages 702--705. IEEE, 2019.

\bibitem{he2020spotting}
Ying He, Su-Jing Wang, Jingting Li, and Moi~Hoon Yap.
\newblock Spotting macro-and micro-expression intervals in long video
  sequences.
\newblock In {\em 2020 15th IEEE International Conference on Automatic Face and
  Gesture Recognition (FG 2020)}, pages 742--748. IEEE, 2020.

\bibitem{zhang2020spatio}
L-w Zhang, Jingting Li, S~Wang, X~Duan, W~Yan, H~Xie, and S~Huang.
\newblock Spatio-temporal fusion for macro-and micro-expression spotting in
  long video sequences.
\newblock In {\em 2020 15th IEEE International Conference on Automatic Face and
  Gesture Recognition (FG 2020)(FG)}, pages 245--252, 2020.

\bibitem{sen2018approximate}
Sanchari Sen and Anand Raghunathan.
\newblock Approximate computing for long short term memory (lstm) neural
  networks.
\newblock {\em IEEE Transactions on Computer-Aided Design of Integrated
  Circuits and Systems}, 37(11):2266--2276, 2018.

\bibitem{bertero1988ill}
Mario Bertero, Tomaso~A Poggio, and Vincent Torre.
\newblock Ill-posed problems in early vision.
\newblock {\em Proceedings of the IEEE}, 76(8):869--889, 1988.

\bibitem{turaga2010advances}
Pavan Turaga, Rama Chellappa, and Ashok Veeraraghavan.
\newblock Advances in video-based human activity analysis: challenges and
  approaches.
\newblock In {\em Advances in Computers}, volume~80, pages 237--290. Elsevier,
  2010.

\bibitem{baltrusaitis2018openface}
Tadas Baltrusaitis, Amir Zadeh, Yao~Chong Lim, and Louis-Philippe Morency.
\newblock Openface 2.0: Facial behavior analysis toolkit.
\newblock In {\em 2018 13th IEEE International Conference on Automatic Face \&
  Gesture Recognition (FG 2018)}, pages 59--66. IEEE, 2018.

\bibitem{zadeh2017convolutional}
Amir Zadeh, Yao Chong~Lim, Tadas Baltrusaitis, and Louis-Philippe Morency.
\newblock Convolutional experts constrained local model for 3d facial landmark
  detection.
\newblock In {\em Proceedings of the IEEE International Conference on Computer
  Vision Workshops}, pages 2519--2528, 2017.

\bibitem{jarrett2009best}
Kevin Jarrett, Koray Kavukcuoglu, Marc'Aurelio Ranzato, and Yann LeCun.
\newblock What is the best multi-stage architecture for object recognition?
\newblock In {\em 2009 IEEE 12th international conference on computer vision},
  pages 2146--2153. IEEE, 2009.

\bibitem{lyu2008nonlinear}
Siwei Lyu and Eero~P Simoncelli.
\newblock Nonlinear image representation using divisive normalization.
\newblock In {\em 2008 IEEE Conference on Computer Vision and Pattern
  Recognition}, pages 1--8. IEEE, 2008.

\bibitem{moilanen2014spotting}
Antti Moilanen, Guoying Zhao, and Matti Pietikainen.
\newblock Spotting rapid facial movements from videos using appearance-based
  feature difference analysis.
\newblock In {\em Pattern Recognition (ICPR), 2014 22nd International
  Conference on}, pages 1722--1727, Aug 2014.

\bibitem{howard2017mobilenets}
Andrew~G Howard, Menglong Zhu, Bo~Chen, Dmitry Kalenichenko, Weijun Wang,
  Tobias Weyand, Marco Andreetto, and Hartwig Adam.
\newblock Mobilenets: Efficient convolutional neural networks for mobile vision
  applications.
\newblock {\em arXiv preprint arXiv:1704.04861}, 2017.

\bibitem{he2016deep}
Kaiming He, Xiangyu Zhang, Shaoqing Ren, and Jian Sun.
\newblock Deep residual learning for image recognition.
\newblock In {\em Proceedings of the IEEE conference on computer vision and
  pattern recognition}, pages 770--778, 2016.

\bibitem{li2022casme3}
Jingting Li, Zizhao Dong, Shaoyuan Lu, Su-Jing Wang, Wen-Jing Yan, Yinhuan Ma,
  Ye~Liu, Changbing Huang, and Xiaolan Fu.
\newblock {{CAS(ME)$^{3}$}: A Third Generation Facial Spontaneous
  Micro-Expression Database with Depth Information and High Ecological
  Validity}.
\newblock {\em IEEE Transactions on Pattern Analysis and Machine Intelligence},
  2022.

\bibitem{li2020megc2020}
LI~Jingting, Su-Jing Wang, Moi~Hoon Yap, John See, Xiaopeng Hong, and Xiaobai
  Li.
\newblock Megc2020-the third facial micro-expression grand challenge.
\newblock In {\em 2020 15th IEEE International Conference on Automatic Face and
  Gesture Recognition (FG 2020)}, pages 777--780. IEEE, 2020.

\bibitem{butterworth1930theory}
Stephen Butterworth et~al.
\newblock On the theory of filter amplifiers.
\newblock {\em Wireless Engineer}, 7(6):536--541, 1930.

\bibitem{pan2020local}
Hang Pan, Lun Xie, and Zhiliang Wang.
\newblock Local bilinear convolutional neural network for spotting macro- and
  micro-expression intervals in long video sequences.
\newblock In {\em 2020 15th IEEE International Conference on Automatic Face and
  Gesture Recognition (FG 2020)(FG)}, pages 343--347, 2020.

\end{thebibliography}

\end{document}